\documentclass[letterpaper, 10pt, conference]{ieeeconf}
\IEEEoverridecommandlockouts

\usepackage{cite}
\usepackage{amsmath}

\usepackage{amsthm}

\usepackage{amssymb}
\usepackage{algorithm}
\usepackage{algpseudocode}
\usepackage{tabularx}
\usepackage{flushend}

\algblockdefx %
[Repeat]{BeginRepeat}{EndRepeat} %
[1][Unknown]{\textbf{repeat} #1 \textbf{times}}%
{\textbf{end repeat}}

\usepackage[caption=false,font=footnotesize]{subfi
g}
\usepackage{hyperref}

\usepackage{fixltx2e}
\usepackage{url}
\usepackage{graphicx}
\usepackage{xcolor}
\definecolor{darkgreen}{HTML}{65C148}
\definecolor{mypurple}{HTML}{C548A4}
\definecolor{darkblue}{HTML}{295DB1}
\definecolor{lightred}{HTML}{FF2E4C}
\definecolor{cyan}{HTML}{17FFFF}
\definecolor{mypink}{HTML}{FF17FF}
\definecolor{purple}{HTML}{6A0DAD}
\usepackage{multirow}
\usepackage{siunitx}

\usepackage{tikz}
\usetikzlibrary{shapes,arrows}
\tikzstyle{block} = [rectangle, fill, fill=blue!10, 
    text width=12em, text centered, minimum height=2em]
\tikzstyle{empty} = [rectangle, fill, fill=white, 
     text centered]
\tikzstyle{line} = [draw, -latex']

\usetikzlibrary{calc}
\newcommand{\normtwo}[1]{\left\lVert#1\right\rVert_2}

\newcommand{\mR}{\mathcal{R}}
\newcommand{\mO}{\mathcal{O}}
\newcommand{\mH}{\mathcal{H}}
\newcommand{\mW}{\mathcal{W}}

\newcommand{\mS}{\mathcal{S}}

\newcommand{\mPxy}[2]{\mathcal{P}_{#1}^{#2}}

\newcommand{\mQ}{\mathcal{Q}}

\newcommand{\vp}{\mathbf{p}}
\newcommand{\vpxy}[2]{\mathbf{p}_{#1}^{#2}}
\newcommand{\vo}{\mathbf{o}}
\newcommand{\vox}[1]{\mathbf{o}_{#1}}
\newcommand{\vf}{\mathbf{f}}
\newcommand{\vfx}[1]{\mathbf{f}_{#1}}
\newcommand{\vfxy}[2]{\mathbf{f}_{#1}^{#2}}

\newcommand{\vn}{\mathbf{n}}
\newcommand{\vgammai}{\boldsymbol{\gamma_i}}
\newcommand{\vx}{\mathbf{x}}

\newcommand{\vhx}[1]{\mathbf{h}_{#1}}

\newcommand{\vlambdai}{\boldsymbol{\lambda}_i}
\newcommand{\vthetai}{\boldsymbol{\theta}_i}
\newcommand{\vPx}[1]{\mathbf{P}_{#1}}

\newcommand{\vgxy}[2]{\mathbf{g}_{#1}^{#2}}

\newcommand{\vPxyzt}[4]{\mathbf{P}_{#1, #2, #3}^{#4}}
\newcommand{\vexyz}[3]{\mathbf{e}_{#1, #2}^{#3}}

\usepackage{pgfplots}
\pgfplotsset{width=7cm,compat=1.8}

\usepackage{array}
\newcolumntype{L}{>{\arraybackslash}m{4cm}}

\usepackage{cleveref}
\begin{document}
\title{RLSS: Real-time Multi-Robot Trajectory Replanning using Linear Spatial Separations
}

\author{Baskın~Şenbaşlar,
        Wolfgang~Hönig, and
        Nora~Ayanian%
\thanks{Baskın~Şenbaşlar and Nora~Ayanian are with the Department of Computer Science, University of Southern California, Los Angeles, CA, USA.}
\thanks{Email: \{baskin.senbaslar, ayanian\}@usc.edu}
\thanks{Wolfgang~Hönig is with the Department of Aerospace, California Institute of Technology, Pasadena, CA, USA. Part of this work was done while he was at the Department of Computer Science, University of Southern California, Los Angeles, CA, USA. Email: whoenig@caltech.edu}
\thanks{This work was supported by NSF awards IIS-1724399, IIS-1724392, and CPS-1837779. B. \c{S}enba\c{s}lar was supported by a USC Annenberg Fellowship.}
}

\maketitle

\begin{abstract}
Trajectory replanning is a critical problem for multi-robot teams  navigating dynamic environments. We present RLSS (Replanning using Linear Spatial Separations): a real-time trajectory replanning algorithm for cooperative multi-robot teams that uses linear spatial separations to enforce safety. Our algorithm handles the dynamic limits of the robots explicitly, is completely distributed, and is robust to environment changes, robot failures, and trajectory tracking errors. It requires no communication between robots and relies instead on local relative measurements only. We demonstrate that the algorithm works in real-time both in simulations and in experiments using physical robots. We compare our algorithm to a state-of-the-art online trajectory generation algorithm based on model predictive control, and show that our algorithm results in significantly fewer collisions in highly constrained environments, and effectively avoids deadlocks.
\end{abstract}


\section{Introduction}

Effective collaboration of multiple robots is key to emerging industries such as warehouse automation~\cite{kiva}, autonomous driving~\cite{autdrive}, and automated intersection management~\cite{intman}.
One of the core robotic challenges in such domains is multi-robot trajectory planning, which entails navigating a team of robots in environments with dynamic obstacles.
In practice, robots must operate safely in dynamic environments even if there is only partial observability and limited communication available.
While there is a large body of research on multi-robot trajectory planning, none of the existing solutions are practical for dynamic environments with high robot- and obstacle-density, which makes the problem particularly challenging.

Existing algorithms employ centralized~\cite{sharon2015,stuckey2019,tang2016safe,honig2018} or distributed strategies~\cite{alonso2013optimal,wang2017safety,primal,bvc,luis2020online} to address the multi-robot trajectory planning problem.
Centralized algorithms can provide theoretical guarantees, but they cannot react to changes in the environment in real-time and they require a reliable communication link to all robots.
Robots using them do not deadlock, and follow trajectories successfully so long as the underlying assumptions hold.
Distributed algorithms delegate the computation of trajectories:
each robot plans for itself and reacts to changes in the environment in real-time.
We propose an algorithm that employs real-time trajectory optimization using receding horizon planning that is fully distributed while exhibiting some advantages typical of centralized solutions that existing distributed algorithms do not have, yet avoiding the shortcomings of existing distributed algorithms that make them impractical.
To our knowledge, compared to other state-of-the-art distributed multi-robot trajectory planning algorithms~\cite{alonso2013optimal,wang2017safety,primal,bvc,luis2020online}, our approach is the only one that provides the following simultaneously. It i) explicitly handles the dynamic limits of the robots; ii) is completely distributed with no reliance on a central computer; iii) is robust to environmental changes, agent failures, and poorly performing controllers; iv) requires no communication between robots; v) does not deadlock; vi) enforces hard safety constraints and reports failure when the constraints are infeasible, thus guaranteeing collision avoidance for computed trajectories; and vii) works in the presence of obstacles. 
Our approach uses a combination of discrete planning and trajectory optimization presented as a convex quadratic program that can be solved in real-time.

\begin{figure}
    \centering
    \includegraphics[width=0.9\linewidth]{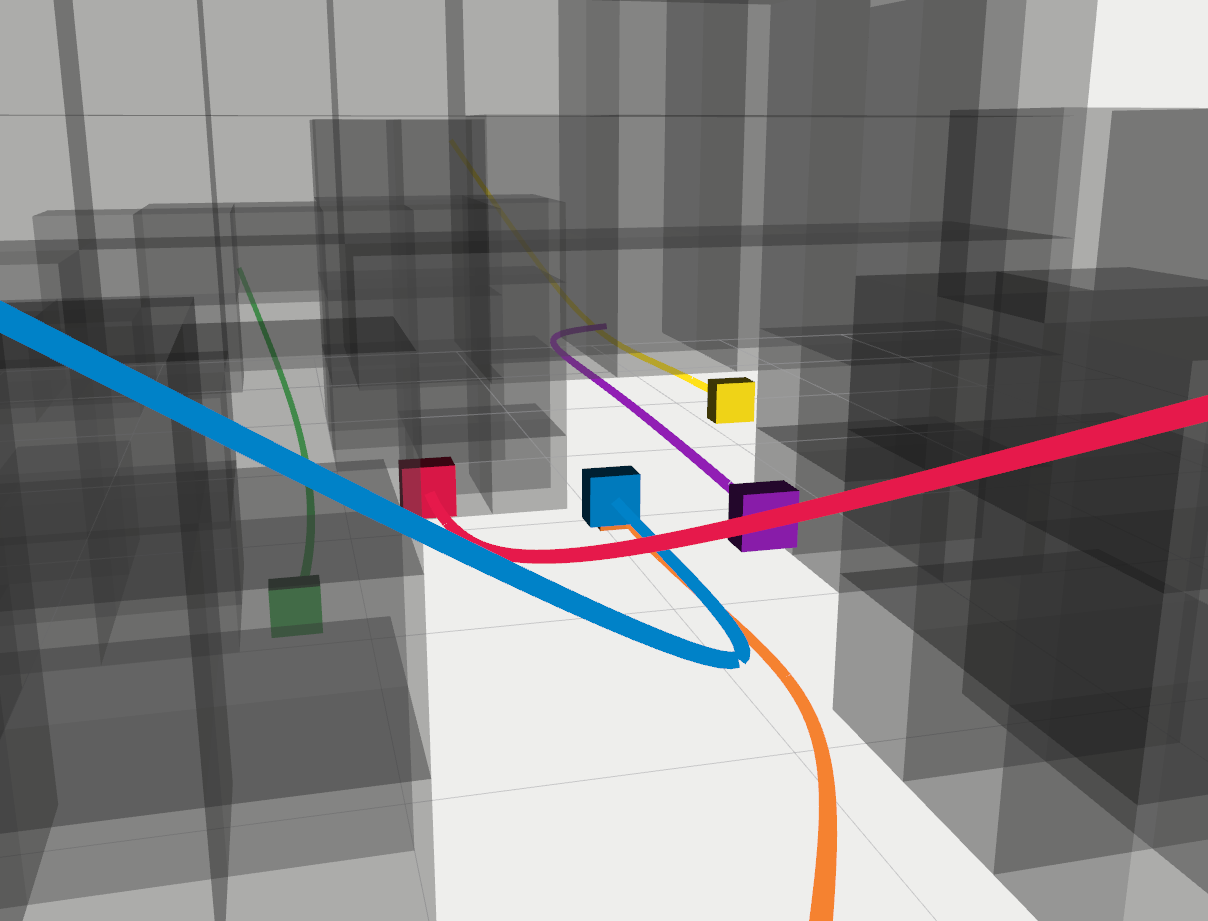}
    \caption{RLSS calculates trajectories in real-time where robots move in close proximity to each other. Computations are done in a distributed way that each robot plans for its own trajectory.}
    \label{fig:nice}
\end{figure}

We utilize linear spatial separations for collision avoidance during the optimization step, hence we call our algorithm RLSS (Replanning using Linear Spatial Separations).
RLSS can replan trajectories for robots in dense environments (Fig.~\ref{fig:nice}) where robots have to move in close proximity to each other and navigate through an environment with hundreds of obstacles.
Each robot plans a trajectory for itself without any reliance on the trajectories of other robots.

We show using simulations that RLSS works in real-time in highly constrained environments.
We compare our algorithm to a state-of-the-art method based on distributed model predictive control (DMPC)~\cite{luis2020online}.
DMPC requires the approximation of the robot's own controller behaviour under given desired states as a linear system, and uses it during planning.
Our method results in significantly fewer collisions and no deadlocks compared to DMPC.
Furthermore, our algorithm does not require communication to ensure safety unlike the baseline algorithm.
We demonstrate our algorithm's robustness to external disturbances in physical robot experiments using a heterogeneous differential drive robot team and a homogeneous quadrotor team.

\section{Problem Statement}

Consider a ---possibly heterogeneous---team of $N$ robots.
We assume the robots are differentially flat~\cite{murray1995differential}, i.e., the robots' states and inputs can be expressed in terms of output trajectories and a finite number of derivatives thereof.
Differential flatness is common for many kinds of mobile robots, including differential drive robots~\cite{campion1996structural}, car-like robots~\cite{murray-car-like}, omnidirectional robots~\cite{omnidirectional}, and quadrotors~\cite{mellinger}.

To provide the collision avoidance guarantee for the computed trajectories, we assume that robots are able to sense the positions of other robots and obstacles around them perfectly.
We assume that obstacles have convex shapes\footnote{For the purposes of our algorithm, concave obstacles can be described or approximated by a union of finite number of convex shapes provided that the union contains the original obstacle. Using our algorithm with approximations of concave obstacles results in trajectories that avoid the approximations.}. Many existing, efficient, and widely-used mapping tools, including occupancy grids~\cite{homm2010efficient} and octomaps~\cite{hornung2013octomap}, internally store obstacles as convex shapes;
such maps can be updated in real-time using visual or RGBD sensors, and 
use unions of convex axis aligned boxes to approximate the obstacles in the environment.
We use $\mO(t)$ to denote the set of obstacles in the environment at time $t$.

Optionally, we permit that each robot $i$ has a desired trajectory $\vo_i(t): [0, T_i] \to \mathbb R^d$, $d\in\{2,3\}$ that it should follow. 
We do not require that this trajectory is collision-free or even feasible to track by the robot. 
If no such desired trajectory is known, it can be initialized, for example, with a straight line to the goal location.

     

Our goal is that each robot $i$ computes and tracks a Euclidean trajectory $\vfx{i}(t) : [0, T_i]\to \mathbb R^d$ such that $\vfx{i}(t)$ is collision-free, executable according to the robot's dynamics, and is as close as possible to the desired trajectory $\vox{i}(t)$.
Since our robots are differentially-flat, we can account for their dynamics by i) imposing $C^c$ continuity on the trajectories for any given $c$, and ii) imposing constraints on the maximum $k^{th}$ derivative magnitude of trajectories $\vfx{i}(t)$ for any desired $k$.





Let $\mS_{R_i}(\vp)$ be the convex collision shape of robot $i$ located at position $\vp \in \mathbb R^d$, i.e. the convex region occupied by robot $i$ located at position $\vp$. 
Let $\mW_i \subseteq \mathbb R^d$ be the convex workspace that robot $i$ should stay inside.
It can be set to a bounding box that defines a room that a ground robot must not leave, a half-space that contains vectors with positive $z$ coordinates so that a quadrotor does not hit the ground or be simply set to $\mathbb R^d$. 
Formally, each robot must solve the following optimization problem:
\begin{equation}
\label{eq:problem}
\begin{aligned}
    \min &\ \ \int_{0}^{T_i} \normtwo{\vfx{i}(t) - \vox{i}(t)}^2 dt \text{ s.t.}\\
    &\vfx{i}(t) \in C^{c_i}\\
    &\frac{d^c\vfx{i}(0)}{dt^c} = \frac{d^c\vpxy{i}{0}}{dt^c} &\forall c\in\{0, \ldots, c_i\}\\
     &\mS_{R_i}(\vfx{i}(t)) \cap \left( \cup_{\mQ\in \mO(t)} \mQ \right) = \emptyset &\forall t\in[0, T_i]\\
     &\mS_{R_i}(\vfx{i}(t)) \cap \mS_{R_j}(\vfx{j}(t)) = \emptyset &\forall j \neq i\; \forall t\in[0,T_i]\\
     &\mS_{R_i}(\vfx{i}(t)) \in \mW_i &\forall t\in[0, T_i]\\
    &\underset{t \in [0, T_i]}{\max}\normtwo{\frac{d^k \vfx{i}(t)}{dt^k}} \leq  \gamma_i^k &\forall k \in \{1, \ldots, K_i\}
\end{aligned}
\end{equation}
where $\vpxy{i}{0}$ is the initial position of robot $i$; $\gamma_i^k$ is the maximum $k^{th}$ derivative magnitude that the $i^{th}$ robot can execute; $K_i$ is the maximum derivative degree that robot $i$ has a derivative magnitude limit on; and $c_i$ is the order of derivative up to which the trajectory of the $i^{th}$ robot must be continuous.

The cost of the optimization problem denotes the deviation from the desired trajectory;
it is the sum of position differences between the planned and the desired trajectories.

\section{Preliminaries}

Here, we introduce essential mathematical concepts we use.

\subsection{Parametric Curves and Splines}
We use curves $\vf : [0, T]\rightarrow \mathbb R^d$ that are parametrized by time, where $T$ is the duration of the curve, as trajectories. 
Mathematically, we adopt splines, i.e. piecewise polynomials, where each piece is a B\'ezier curve defined by a set of control points and a duration. 

A B\'ezier curve $\vf : [0, T] \rightarrow \mathbb R^d$ of degree $h$ is defined by $h+1$ control points $\vPx{0}, \ldots, \vPx{h} \in \mathbb R^d$ as follows:
\begin{equation}
\begin{aligned}
    \vf(t) &= \sum_{i=0}^h \vPx{i} {h \choose i} \left(\frac{t}{T}\right)^i\left(1-\frac{t}{T}\right)^{(h-i)}.
\end{aligned}
\label{bezier-definition}
\end{equation}
Since any B\'ezier curve $\vf$ is a polynomial of degree $h$, it is smooth, meaning $\vf\in C^\infty$.

We focus on B\'ezier curves as pieces because of their \emph{convex hull property}: the curves themselves lie inside the convex hull of their control points, i.e., $\vf(t) \in ConvexHull\{\vPx{0},\ldots,\vPx{h}\}\ \forall t\in[0, T]$~\cite{Bernstein}.
Using the convex hull property, we can constrain a curve to be inside a convex region by constraining its control points to be inside the same convex region.

\subsection{Linear Spatial Separations: Half-spaces, Convex Polytopes, and Support Vector Machines}

A hyperplane $\mH$ in $\mathbb R^d$ can be formulated by a normal vector $\vn$ and an offset $a$ as $\mH = \{\vx\in \mathbb R^d\ |\ \vn^\top\vx + a = 0\}$. 
A half-space $\tilde{\mH}$ in $\mathbb R^d$ is a subset of $\mathbb R^d$ that is bounded by a hyperplane such that $\tilde{\mH} = \{\vx\in \mathbb R^d\ |\ \vn^\top \vx + a \leq 0\}$.
A convex polytope is the intersection of a finite number of half-spaces.

Our approach relies heavily on computing safe convex polytopes and constraining spline pieces to be inside these polytopes.
Specifically, we compute support vector machine (SVM)~\cite{cortes1995support} hyperplanes between curve pieces and the obstacles/robots in the environment, and use these hyperplanes to create safe convex polytopes for robots to navigate in. 



\section{Approach} \label{section:approachoverview}
To solve~\eqref{eq:problem} in one shot, it is required for the robot to know the trajectories $\vfx{j}(t)$ of the other robots and the future obstacle positions $\mO(t)$ ahead of time. While the first requirement could be realized by some form of inter-robot communication, the second requirement is unrealistic for dynamic environments. Therefore, instead of solving this problem exactly in one shot, we solve it iteratively by replanning at each timestep.

Each robot $i \in \{1, \ldots, N\}$ replans at every timestep $u$ to calculate a piecewise trajectory $\vfxy{i}{u}(t)$ that is safe for duration $\delta t$, where each piece is a B\'ezier curve. 
Then, it executes $\vfxy{i}{u}(t)$ for $\delta t$ and replans.
Henceforth, we will use superscript $u$  to denote an object at timestep $u$ and subscript $i$ to denote an object used or computed by robot $i$.

RLSS fits into the planning part of the classical robotics pipeline using perception, planning, and control. 
The inputs from the perception for each robot $i$ at each timestep $u$ are:
\begin{itemize}
    \item $\mR_i^u$: Convex collision shapes of other robots\footnote{Practically, if robot $i$ cannot sense a particular robot $j$ at timestep $u$ because it is not within the sensing range of $i$, robot $j$ can be omitted by robot $i$.
    As long as the sensing range of robot $i$ is more than the maximum distance that can be travelled by robots $i$ and $j$ in duration $\delta t$, omitting robot $j$ does not affect the safety of the algorithm.}. $\mR_{i,j}^u \in \mR_i^u$ where $j \in \{1,\ldots, i-1, i+1,\ldots,N\}$ is the convex collision shape of robot $j$ sensed by robot $i$ at timestep $u$.
    \item $\mO_i^u$: The set of convex obstacles robot $i$ sensed up to timestep $u$.
    \item $\vpxy{i}{u}$: Position of robot $i$ at timestep $u$ estimated by itself.
\end{itemize}

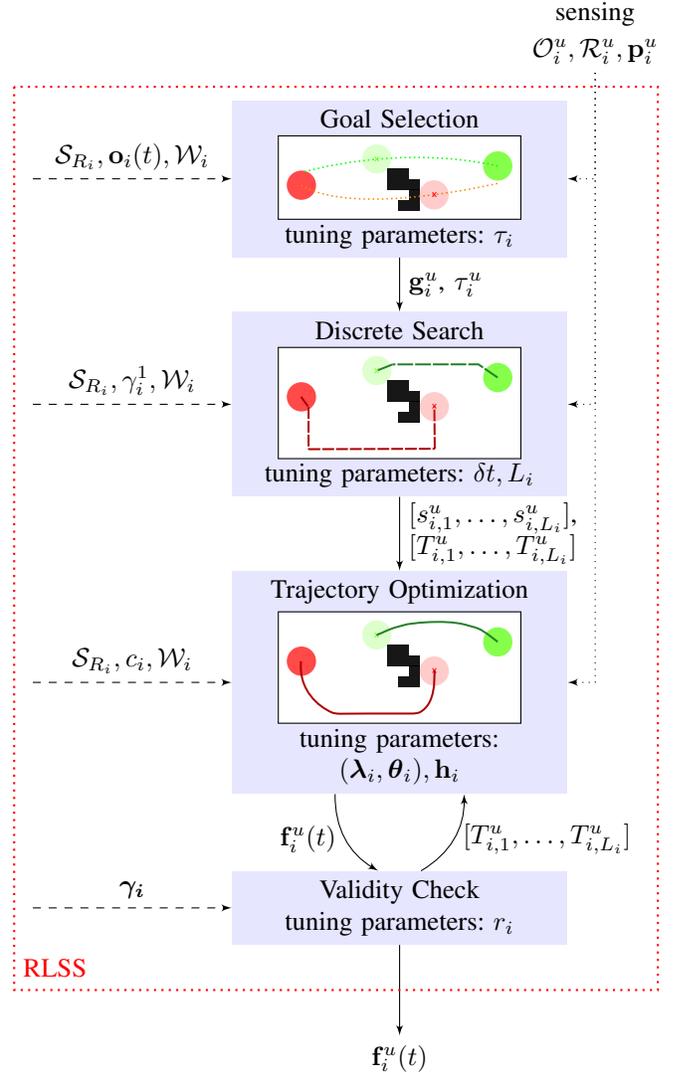
\begin{figure}
\centering
\begin{tikzpicture}
    \node [empty] (goalselectioninput) {};
    \node [empty, below of = goalselectioninput, node distance = 3cm] (discretesearchinput) {};
    \node [empty, below of = goalselectioninput, node distance = 6.7cm] (trajoptinput) {};
    \node [empty, below of = trajoptinput, node distance = 3cm] (tempresinput) {};
    \node [block, right of = goalselectioninput,  node distance = 5cm] (goalselection) {Goal Selection\\\vspace*{0.1cm} \fcolorbox{black}{white}{\includegraphics[width=3cm]{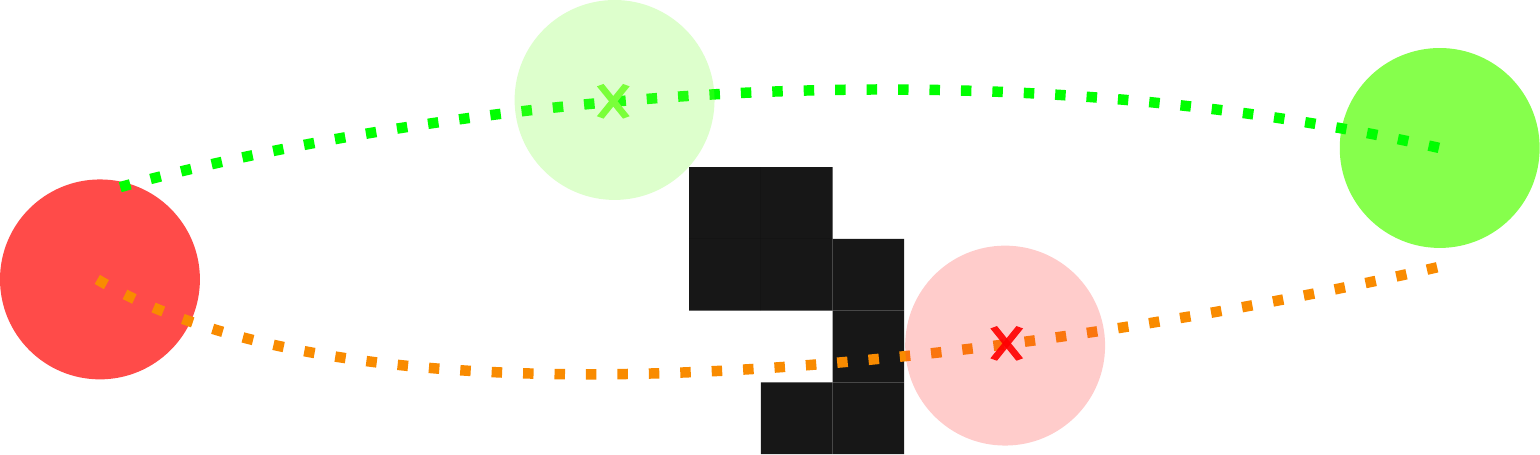}}\\tuning parameters: $\tau_i$};
    \node [block, below of = goalselection, node distance=3cm] (discretesearch) {Discrete Search\\\vspace*{0.1cm} \fcolorbox{black}{white}{\includegraphics[width=3cm]{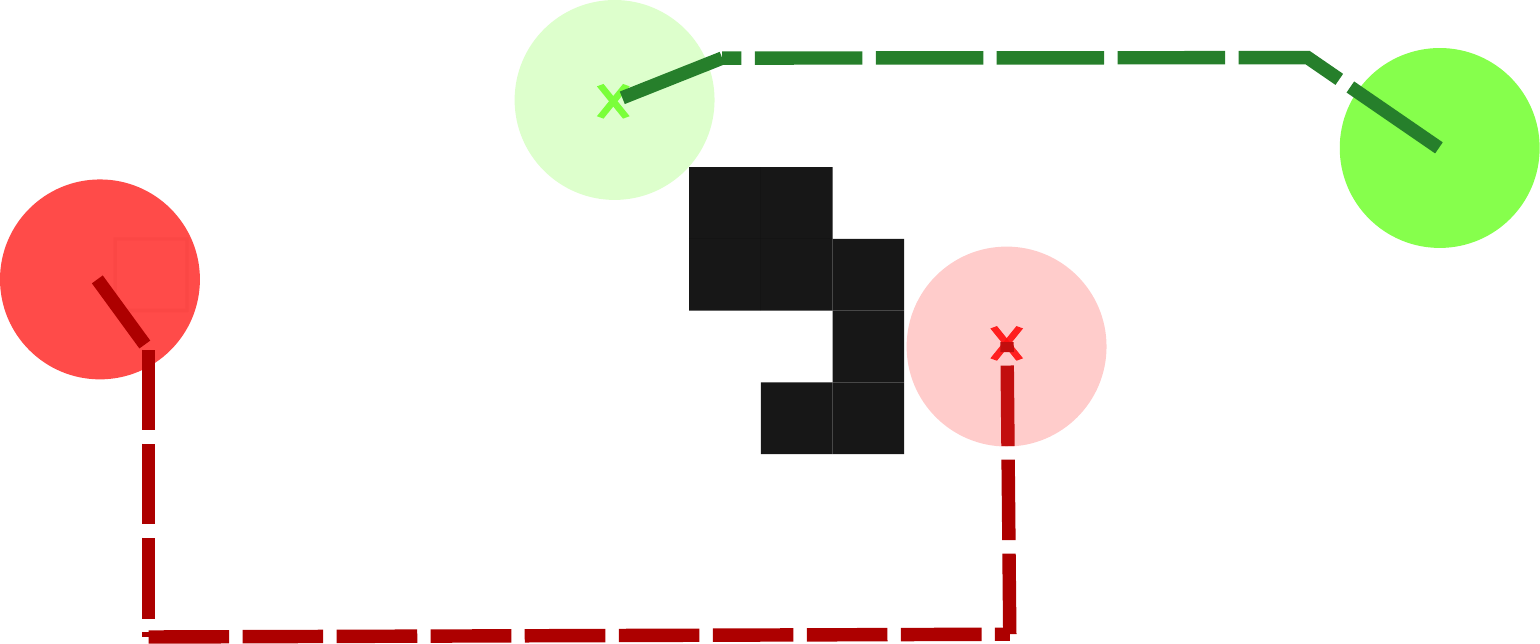}}\\tuning parameters: $\delta t, L_i$};
    \node [block, below of = discretesearch, node distance=3.7cm] (trajopt) {Trajectory Optimization\\\vspace*{0.1cm} \fcolorbox{black}{white}{\includegraphics[width=3cm]{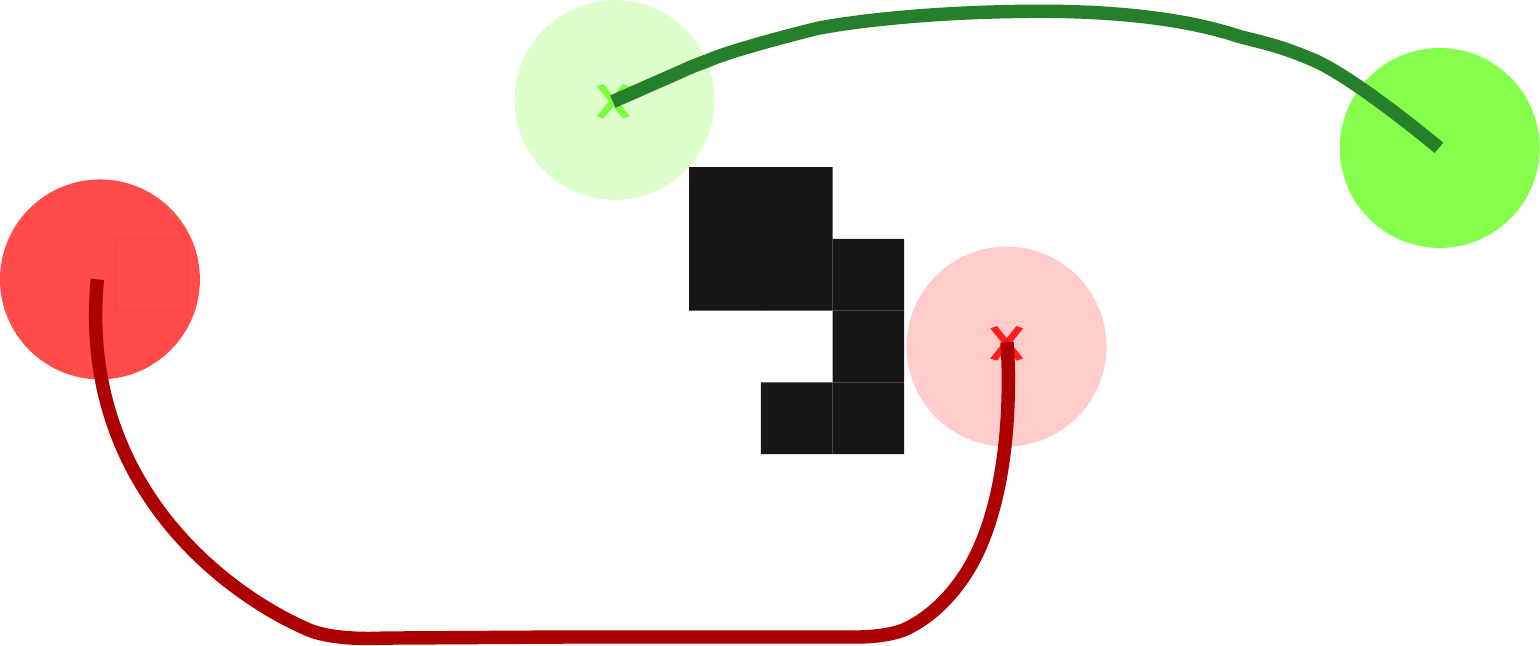}}\\tuning parameters: $(\vlambdai, \vthetai), \vhx{i}$};
    \node [block, below of = trajopt, node distance=3cm] (tempres) {Validity Check\\tuning parameters: $r_i$};
    \node [empty, below of = tempres, node distance = 2cm] (end) {$\vfxy{i}{u}(t)$};
    \node [empty, above of = goalselection, node distance = 2.2cm] (sense1) {};
    \node [empty, right of = sense1, node distance = 2.6cm, align=center] (sense) {sensing};
    \node [empty, below of = sense, node distance = 0.5cm] (obstacles) {$\mO_i^u, \mR_i^u, \vpxy{i}{u}$};
    \node[text width=4cm, color = red] at (2,-10.5) {RLSS};
    
    \path [line] (goalselection) -- node [right] {$\vgxy{i}{u}$, $\tau_i^u$} (discretesearch);
    \path [line] (discretesearch) -- node [right, align=left] {$[s_{i, 1}^u, \ldots, s_{i,L_i}^u]$,\\ $[T_{i,1}^u,\ldots, T_{i,L_i}^u]$} (trajopt);
    \draw [-latex'] (trajopt) to[bend right] node [left] {$\vfxy{i}{u}(t)$} (tempres);
    \draw [-latex'] (tempres) to[bend right] node [right] {$[{T}^{u}_{i,1}, \ldots, {T}^{u}_{i, L_i}]$} (trajopt);
    \path [line] (tempres) -- (end);
    \path [line, dashed] (goalselectioninput) -- node [above] {$\mS_{R_i}, \vox{i}(t), \mW_i$} (goalselection);
    \path [line, dashed] (discretesearchinput) -- node [above] {$\mS_{R_i}, \gamma_i^1, \mW_i$} (discretesearch);
    \path [line, dashed] (trajoptinput) -- node [above] {$\mS_{R_i}, c_i, \mW_i$} (trajopt);
    \path [line, dashed] (tempresinput) -- node [above] {$\vgammai$} (tempres);
    \path [line, dotted] (obstacles) |- (goalselection);
    \path [line, dotted] (obstacles) |- (discretesearch);
    \path [line, dotted] (obstacles) |- (trajopt);
    
    \draw[red,thick,dotted] ($(goalselectioninput.north west)+(0,1.1)$)  rectangle ($(tempres.south east)+(1.2,-0.6)$);
\end{tikzpicture}
\caption{The overall data flow pipeline of RLSS. Based on the sensed obstacles $\mO_i^u$, robots $\mR_i^u$, and estimated position $\vpxy{i}{u}$ we compute the trajectory $\vfxy{i}{u}$ at every timestep $u$. The problem instance constants $S_{R_i}, \vox{i}(t), \mW_i, \vgammai$, and $c_i$ are given with dashed lines to stages that use them as inputs. More details about each stage are explained in our journal submission.}
\label{figure:pipeline}
\end{figure}

There are $4$ main stages of RLSS: 1) goal selection, 2) discrete search, 3) trajectory optimization, and 4) validity check.
The replanning pipeline is summarized in Fig.~\ref{figure:pipeline}.
At each timestep $u$, each robot $i$ executes the four stages independently of the other robots.

In the goal selection stage, a goal position $\vgxy{i}{u}$ on the robot's desired trajectory $\vox{i}(t)$ is selected.
Given the desired time horizon $\tau_i$, which denotes the desired duration of the resulting trajectory, we adjust $\tau_i$ to the actual time horizon $\tau_i^u$ so that the position $\vgxy{i}{u} = \vox{i}(u\delta t + \tau_i^u)$ is a reachable position for the robot without considering kinematics. 
The computed goal position $\vgxy{i}{u}$ of robot $i$ at timestep $u$ and the actual time horizon $\tau_i^u$ of robot $i$ at timestep $u$ is given as input to the discrete search.
Details of the goal selection stage are presented in our journal submission.

During the discrete search stage, we compute a discrete path with no collisions within the workspace $\mW_i$ from the robot's current position $\vpxy{i}{u}$ towards its goal position $\vgxy{i}{u}$.
Also, discrete search calculates the total duration $T_i^u$ for the path using $\tau_i^u$, $\delta t$, $\gamma_i^1$, and distributes the total duration to each of the segments in the discrete path proportional to the segment lengths.
The desired number of segments of the discrete path is set by the parameter $L_i$.
The outputs are the resulting path segments $s_{i,1}^u, \ldots, s_{i,L_i}^u$, and path segment durations $T_{i, 1}^u, \ldots, T_{i, L_i}^u$, where $s_{i,j}^u$ is the $j^{th}$ segment of the discrete path for robot $i$ at timestep $u$ and $T_{i,j}^u$ is that segment's duration.
Details of the discrete search stage are presented in our journal submission.

At the trajectory optimization stage, we compute convex polytopes between i) obstacles $\mO_i^u$ and each volume swept by the robot with collision shape function $\mS_{R_i}$ while traversing a discrete path segment, and ii) robot collision shapes $\mR_{i}^u$ and each volume swept by the robot while traversing a discrete path segment using support vector machines. 
The optimization smooths the path segments within these convex polytopes to a piecewise trajectory where each piece is a B\'ezier curve so that the resulting trajectory is safe in terms of both robot-to-robot and robot-to-obstacle collisions.
The smoothing procedure is formulated as a convex quadratic optimization problem.
Let $T_i^u = \sum_{j=1}^{L_i}T_{i, j}^u$ denote the total duration of the planned trajectory where $T_{i, j}$ is the duration assigned to the $j^{th}$ segment. 
The B\'ezier piece degrees, and hence the number of control points of pieces, are tuned with the parameter $\vhx{i}$, where the $j^{th}$ entry of $\vhx{i}$, namely $h_{i,j}$, is the degree of the $j^{th}$ piece of the trajectory for each $j \in \{1, \ldots, L_i\}$.
Let $\vPxyzt{i}{j}{k}{u} \in \mathbb R^d$ be the $k^{th}$ control point of the $j^{th}$ B\'ezier piece of robot $i$ at timestep $u$. Let $\mPxy{i}{u} = \{\vPxyzt{i}{j}{k}{u}\ |\ j \in \{1, \ldots, L_i\}, k \in \{0, \ldots, h_{i, j}\} \}$ be the set of all control points of robot $i$ at timestep $u$.

The cost function we use in the trajectory optimization stage is a weighted combination of energy usage and deviation from the discrete segments. 

We use the sum of integrated squared derivative magnitudes as a metric for energy usage, similar to prior work~\cite{honig2018, richterISRR}. 
The energy usage $E(\mPxy{i}{u})$ is computed as
\begin{equation}
\begin{aligned}
    E(\mPxy{i}{u}) = \sum_{j} \lambda_{i, j} \int_{0}^{T_i^u}\normtwo{\frac{d^j\vfxy{i}{u}(t)}{dt^j}}^2dt,
\end{aligned}
\end{equation}
where $\lambda_{i, j}$ are weight parameters. 
$E(\mPxy{i}{u})$ is a quadratic function of the control points $\mPxy{i}{u}$~\cite{richterISRR}.

We use the Euclidean distance between trajectory piece endpoints and segment endpoints as a metric for the deviation from the discrete segments.
The deviation $D(\mPxy{i}{u})$ from the discrete segments is 
\begin{equation}
\begin{aligned}
    D(\mPxy{i}{u}) = \sum_{j=1}^{L_i} \theta_{i, j} \normtwo{\vPxyzt{i}{j}{h_{i, j}}{u} - \vexyz{i}{j}{u}}^2,
\end{aligned}
\end{equation}
where $\theta_{i, j}$ are the weight parameters, $\vexyz{i}{j}{u}$ is the endpoint of the segment $s_{i, j}^u$, and $\vPxyzt{i}{j}{h_{i, j}}{u}$ is the last control point of the $j^{th}$ piece.

In addition to the convex polytope constraints we impose for safety, we add constraints for differential continuity between pieces, differential continuity between planning iterations and navigation within the workspace.

The output of the trajectory optimization is the resulting trajectory $\vfxy{i}{u}(t)$.

During the validity check, we check i) whether the trajectory optimization succeeded and ii) whether the trajectory obeys the robot's maximum derivative magnitudes $\vgammai = (\gamma_i^1, \ldots, \gamma_i^{K_i})^\top$.
If any of the checks fail, temporal re-scaling is applied.
Each piece duration is multiplied by $r_i$, and trajectory optimization is executed again with the new durations.
This process is repeated until the trajectory passes the validity check.
Details of the validity check stage are presented in our journal submission.

The output of one iteration of RLSS is the final output of the trajectory optimization stage which has duration $\tau_i^u$ and is safe for duration $\delta t$.

We provide no theoretical guarantees on the completeness of RLSS for arbitrary robot dynamics.
Therefore, replanning may fail.
In our usage of the algorithm, we use the trajectory calculated in the previous iteration when replanning fails.
However, one can also choose to stop the robot immediately depending on the robot dynamics.

\section{Experiments} \label{section:experiments}
We implement and release an implementation of our algorithm in C++\footnote{\url{https://github.com/usc-actlab/rlss}} with its ROS integration\footnote{\url{https://github.com/usc-actlab/rlss\_ros}}. 
Our implementation of the algorithm requires the occupancy grid representation of the environment.
It regards each occupied cell as a separate obstacle. 

We test our algorithm in simulations and on physical robots. 
During each type of experiment, we feed the pre-initialized occupancy grid of the environment directly to the algorithm and we do not integrate a mapping system to the experiments.

A supplemental video for our experiments is available at~\url{ https://youtu.be/xsnWs-85knA}.

\subsection{Experiments on Simulated Robots}
We simulate our algorithm on a desktop computer (Intel i9-9900R @ 3.10GHz CPU, 32GB memory) with Ubuntu 18.04 LTS as the operating system.
We use CPLEX~\cite{cplex2009v12} to solve the trajectory optimization problem in our implementation.

\begin{table}[]
    \centering
    \caption{Descriptions of Setups}
    \label{table:setups}
    \begin{tabular}{|c|c|c|c|c|c|}
        \hline
         \# & Dim. & \# Robots & Cell Size & \# Cells & \# Obstacles \\
        \hline
         1 & 3D & 6 & 50x50x50 $\si{cm}^3$ & 32k & 114\\
        \hline
         2 & 3D & 6 & 40x40x40 $\si{cm}^3$ & 62k & 172\\
        \hline
         3 & 3D & 6 & 30x30x30 $\si{cm}^3$ & 148k & 246\\
        \hline
         4 & 3D & 6 & 20x20x20 $\si{cm}^3$ & 500k & 601\\
        \hline
         5 & 3D & 6 & 10x10x10 $\si{cm}^3$ & 4m & 2736\\
        \hline
        7 & 3D & 12 & 50x50x50 $\si{cm}^3$ & 32k & 114\\
        \hline
        8 & 3D & 24 & 50x50x50 $\si{cm}^3$ & 32k & 114\\
        \hline
        9 & 3D & 48 & 50x50x50 $\si{cm}^3$ & 32k & 114\\
        \hline
        10 & 3D & 6 & 50x50x50 $\si{cm}^3$ & 32k & 0\\
        \hline
        11 & 3D & 12 & 50x50x50 $\si{cm}^3$ & 32k & 0\\
        \hline
        12 & 3D & 24 & 50x50x50 $\si{cm}^3$ & 32k & 0\\
        \hline
        13 & 3D & 48 & 50x50x50 $\si{cm}^3$ & 32k & 0\\
        \hline
    \end{tabular}
\end{table}

\begin{figure}
    \centering
     \subfloat[Desired Trajectories\label{figure:cellsize_tests_a}]{%
       \includegraphics[width=0.48\linewidth]{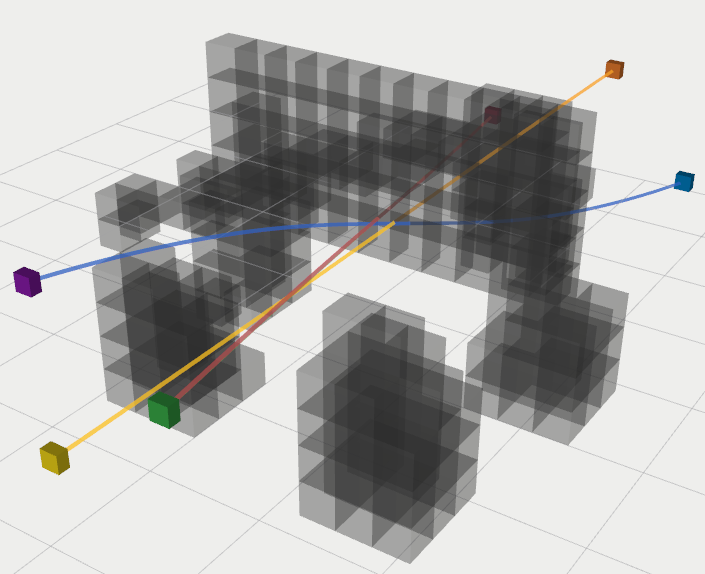}
     }
     \hfill
     \subfloat[Executed Trajectories\label{figure:cellsize_tests_b}]{%
       \includegraphics[width=0.48\linewidth]{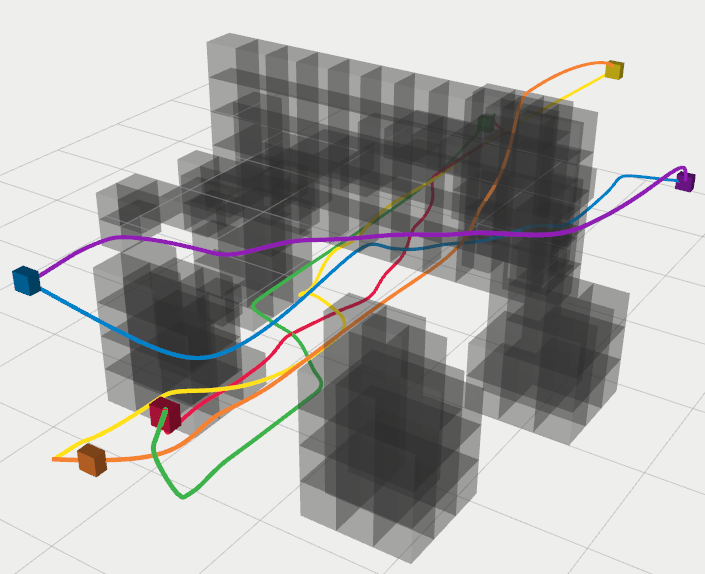}
     }
     
     \caption{The environment and desired trajectories in setups $1,2,3,4,$ and $5$. There are 6 robots that needs to cross a tight environment in opposite directions. Desired trajectories of the robots are given in (a). The executed trajectories in setup $2$ calculated by RLSS in real-time are given in (b).}
     \label{figure:cellsize_tests}
\end{figure}

Descriptions of the setups we use during the simulations are summarized in Table~\ref{table:setups}.
Setups $1$, $2$, $3$, $4$, $5$, $7$, $8$, and $9$ contain the same set of obstacles shown in Fig.~\ref{figure:cellsize_tests} differing in the cell size they use for occupancy grids or the number of robots they contain. 
Setups $10$, $11$, $12$, and $13$ contain no obstacles. They differ in the number of robots navigating in the environment.

All of the experiments presented here are conducted in 3D.
2D experiments exists in our journal submission.
We use two strategies for solving optimization problems in the experiments. 
In our first strategy, which we call HARD, we solve the hard constraint formulation of the optimization problem.
If the hard constraint formulation fails, trajectory optimization stage fails.
In our second strategy, which we call HARD-SOFT, we first solve the hard constraint formulation of the problem and fall back to soft constraint formulation of the problem whenever the hard version fails.
If the soft constraint formulation fails, trajectory optimization stage fails.

\begin{table}[]
    \centering
    \caption{Computation Time per Iteration with Varying Cell Sizes}
    \label{table:cellsize}
    \begin{tabular}{|c|c|c|c|c|c|}
        \cline{3-6}
        \multicolumn{2}{c|}{} & \multicolumn{2}{c|}{HARD} & \multicolumn{2}{c|}{HARD-SOFT}\\
         \hline\rule{0pt}{2ex}
         \# & C. Size [$\si{cm}^3$] & max [\si{ms}] & avg [\si{ms}] & max [\si{ms}] & avg [\si{ms}] \\
         \hline\rule{0pt}{2ex}
         1 & 50x50x50 &88.83 & 13.57 & 88.77 & 13.08\\
         \hline\rule{0pt}{2ex}
         2 & 40x40x40 & 89.31 & 12.34 & 76.19 & 12.32\\
         \hline\rule{0pt}{2ex}
         3 & 30x30x30 & 100.56 & 14.26 & 167.59 & 14.48\\
         \hline\rule{0pt}{2ex}
         4 & 20x20x20 & 103.49 & 19.72 & 120.13 & 19.69\\
         \hline\rule{0pt}{2ex}
         5 & 10x10x10 & 4598.22 & 159.40 & 4597.91 & 159.44\\
         \hline
         
    \end{tabular}
\end{table}

We check how the occupancy grid cell size affects the performance of our planning pipeline.
We use the scenario shown in Fig.~\ref{figure:cellsize_tests} in which 6 robots cross a tight 3D environment.
We conduct our experiments using both HARD and HARD-SOFT strategies.
We plan for $4$-piece splines with degree $6$ B\'ezier curves as pieces.
The results of our experiments are summarized in Table~\ref{table:cellsize}.
Falling back to the soft formulation when the hard formulation fails does not significantly change the computation time per iteration on average as the hard formulation succeeds most of the time.
However, as seen in setups $3$ and $4$, the soft formulation of the optimization problem can decrease the worst case performance of the RLSS pipeline.
While we can run RLSS at more than $\SI{5}{Hz}$ in the worst case in setups $1$, $2$, $3$, and $4$, our algorithm performs poorly for setup $5$ in which the cell size is significantly smaller.

\begin{figure}
    \centering
     \subfloat[Setup 1\label{figure:robotnum_a}]{%
       \includegraphics[width=0.49\linewidth]{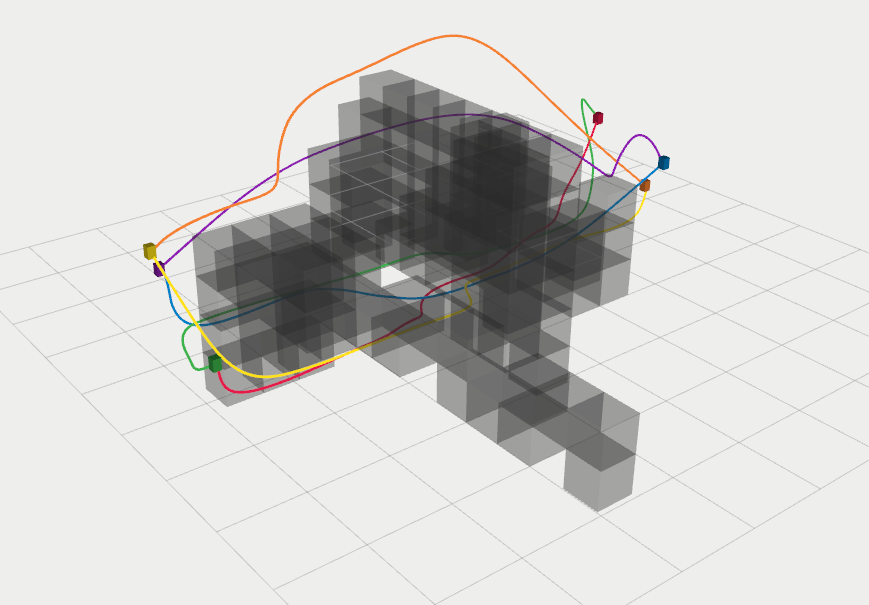}
     }
     \subfloat[Setup 7\label{figure:robotnum_b}]{%
       \includegraphics[width=0.49\linewidth]{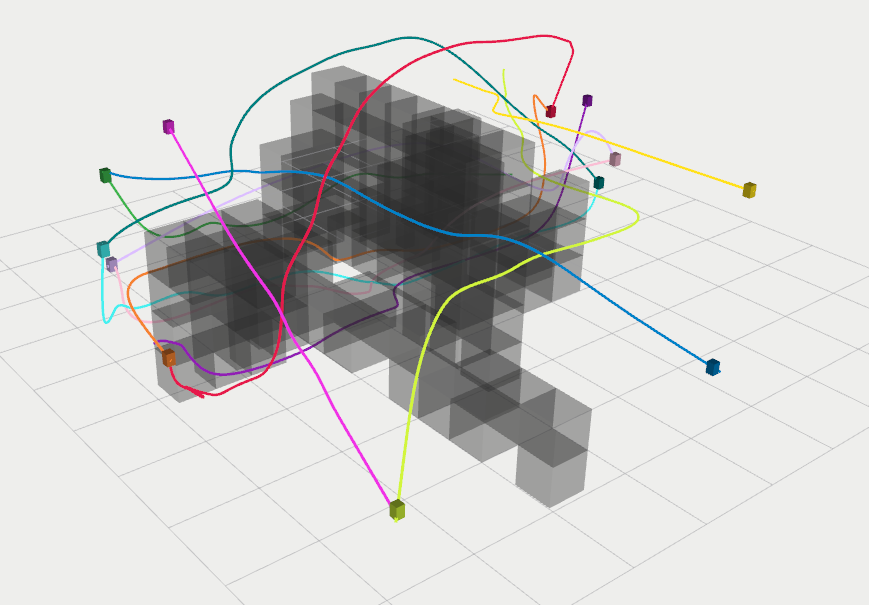}
     }
     \hfill
    \subfloat[Setup 8\label{figure:robotnum_c}]{%
       \includegraphics[width=0.49\linewidth]{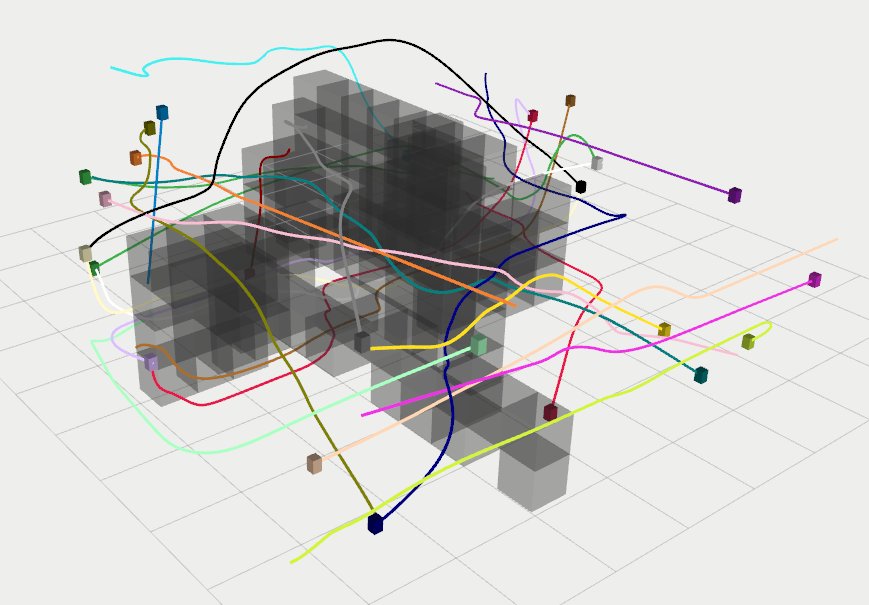}
     }
     \subfloat[Setup 9\label{figure:robotnum_d}]{%
       \includegraphics[width=0.49\linewidth]{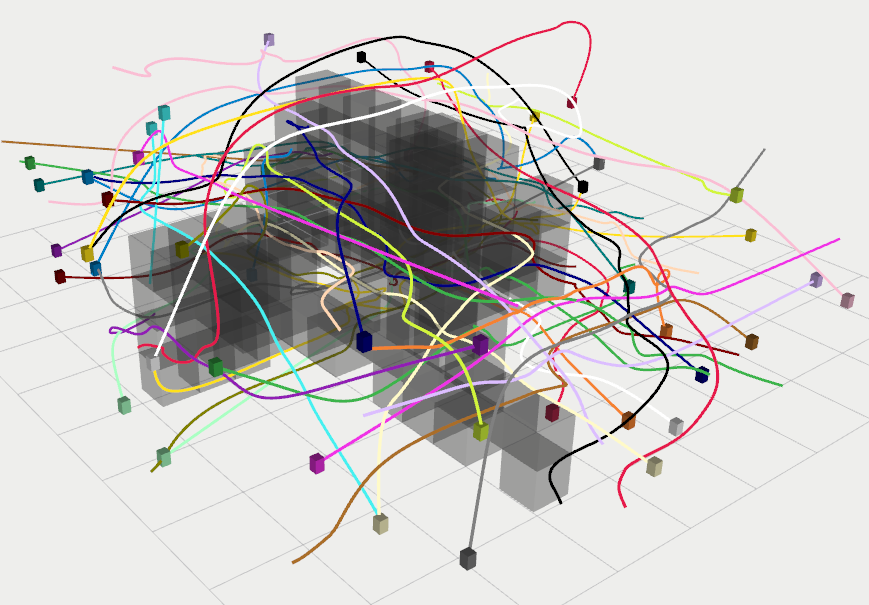}
     }
     
     \caption{Executed trajectories in setups $1$, $7$, $8$, and $9$ when the HARD strategy is used.}
     \label{figure:robotnum}
\end{figure}

Next, we test the scalability with respect to the number of robots. 
We use the environment in Fig.~\ref{figure:cellsize_tests} with 50x50x50 $\si{cm}^3$ cell sizes. 
We test with $6$, $12$, $24$, and $48$ robots corresponding to setups $1$, $7$, $8$, and $9$.
We conduct our experiments using both HARD and HARD-SOFT strategies.
We plan for $4$-piece splines with degree $6$ B\'ezier curves as pieces.
The resulting executed trajectories for the HARD strategy are shown in Fig.~\ref{figure:robotnum}.
The results of our experiments are summarized in Table~\ref{table:robot_scalability_hard}.
As it is the case for varying occupancy grid cell sizes, falling back to soft formulation when the hard formulation fails does not significantly change the performance of our pipeline on average.
However, the worst case performance of RLSS decreases significantly when the soft formulation is introduced.
While we can run our algorithm at $\SI{10}{Hz}$ for experiments $7$ and $8$ with the HARD strategy, we can only run it at $\SI{5}{Hz}$ with the HARD-SOFT strategy in the worst case.
In experiment $9$, the HARD-SOFT strategy takes more than $3$ times more time than the HARD strategy in the worst case.

\begin{table}[]
    \centering
    \caption{Computation Time per Iteration With Varying Number of Robots}
    \label{table:robot_scalability_hard}
    \begin{tabular}{|c|c|c|c|c|c|}
        \cline{3-6}
        \multicolumn{2}{c|}{} & \multicolumn{2}{c|}{HARD} & \multicolumn{2}{c|}{HARD-SOFT}\\
         \hline
         \# & \# Robots & max [$\si{ms}$] & avg [$\si{ms}$] & max [$\si{ms}$] & avg [$\si{ms}$]\\
         \hline
         1 & 6 & 88.83 & 13.57& 88.77 & 13.08\\
         \hline
         7 & 12 & 101.17 & 11.60& 200.78 & 12.60\\
         \hline
         8 & 24 & 105.63 & 11.84& 216.50 & 11.88\\
         \hline
         9 & 48 & 142.38 & 15.94& 532.56 & 18.38\\
         \hline
    \end{tabular}
\end{table}

\subsection{Comparison with DMPC in Simulations}

We compare the performance of our algorithm with a state-of-the-art online trajectory generation algorithm based on distributed model predictive control, which we will call DMPC from now on~\cite{luis2020online}.
DMPC uses a quadratic optimization formulation where safety is enforced with soft constraints.
It requires communication between robots;
robots exchange their intended future states via a communication channel whenever they replan, and
robots receiving the future states add constraints that enforce a minimum distance to the future states of other robots.
Hence, safety depends on the quality of the communication link.
The optimization problem DMPC formulates is goal oriented in which robots try to go as close as possible to their goals in each iteration.
During comparisons, we use the HARD-SOFT strategy in our algorithm because it is the strategy we would employ in a real-life deployment.

There are two sets of setups we use during the comparison.
In the first set, we use setups $10$, $11$, $12$, and $13$ in which there are no obstacles and the number of robots are $6$, $12$, $24$, and $48$, respectively.
In the second set, we use setups $1$, $7$, $8$, and $9$ in which there are $114$ obstacles and the number of robots are $6$, $12$, $24$, and $48$ respectively.

During the comparisons, we use B\'ezier curves of degree $3$ since DMPC optimization fails for larger degrees in its authors' implementation.
We plan for $4$-piece splines with degree $3$ B\'ezier curves as pieces using both algorithms.

First, we compare RLSS and DMPC in terms of computation time per iteration to quantify performance.
The results of computation time comparison are summarized in Table~\ref{tab:runtime_comparison}.
DMPC takes considerably less time both in the worst case and in the average case than RLSS.

\begin{table}[]
    \centering
    \caption{Computation Time per Iteration Comparison of RLSS and DMPC}
    \label{tab:runtime_comparison}
    \begin{tabular}{|c|c|c|c|c|}
        \cline{2-5}
        \multicolumn{1}{c|}{}& \multicolumn{2}{c|}{RLSS} & \multicolumn{2}{c|}{DMPC}\\
        \hline
         \#& max [\si{ms}] & avg [\si{ms}] & max [\si{ms}] & avg [\si{ms}]\\
         \hline 
         10 & 41.02 & 8.25 & \textbf{10.97} & \textbf{1.05}\\
         \hline
         11 & 135.66 & 9.49 & \textbf{11.17} & \textbf{0.99}\\
         \hline
         12 & 165.98 & 11.44 & \textbf{11.31} & \textbf{0.96}\\
         \hline
         13 & 272.33 & 14.78 & \textbf{6.18} & \textbf{1.05}\\
         \hline
         1 & 167.37 & 17.60 & \textbf{13.43} & \textbf{2.55}\\
         \hline
         7 & 213.97 & 14.11 & \textbf{6.66} & \textbf{2.11}\\
         \hline
         8 & 226.79 & 14.07 & \textbf{12.51} & \textbf{1.68}\\
         \hline
         9 & 270.46 & 17.15 & \textbf{7.13} & \textbf{1.64}\\
         \hline
    \end{tabular}
\end{table}

Then, we compare RLSS and DMPC in terms of number of deadlocks, number of collisions, and total distance traveled by all robots to quantify solution quality.
The number of collisions is calculated by sampling the trajectories traversed by robots in $0.01$ second intervals and incrementing a collision counter by $1$ for each robot/robot and robot/obstacle collision.
If there is more than $1$ collision in the same $0.01$ second interval, collision counter is incremented by the number of collisions.
The results of our comparisons are summarized in Table~\ref{tab:quality_comparison}.
In our experiments, robots using RLSS do not deadlock unlike robots that employ DMPC.
Also, in all of the cases, DMPC causes collisions.
This stems from the fact that DMPC uses a soft QP formulation.
On the other hand, the number of collisions with RLSS are considerably lower, because RLSS enforces safety with hard constraints, and falls back to a soft formulation only if the hard version fails.
Lastly, the distance travelled is lower in DMPC than RLSS in $3$ out of $4$ experiments without deadlocks (experiments with setups $10$, $11$, $12$, $13$).
This happens because DMPC utilizes communication to exchange planned trajectories, which results in a more efficient use of the free space.
On the other hand, since RLSS depends on sensing only, robots using it share the available space conservatively.

\begin{table}[]
    \centering
    \caption{Quality Comparison of RLSS and DMPC in Terms of Number of Deadlocks, Number of Collisions and Total Distance Travelled}
    \label{tab:quality_comparison}
    \begin{tabular}{|c|c|c|c|c|c|c|}
        \cline{2-7}
        \multicolumn{1}{c|}{}& \multicolumn{3}{c|}{RLSS} & \multicolumn{3}{c|}{DMPC}\\
        \hline
         \#& \# deadl. & \# coll. & dist. [\si{m}] & \# deadl. & \# coll. & dist. [\si{m}]\\
         \hline 
         10 & \textbf{0}& \textbf{0} & 44.31 & \textbf{0} & 33 & \textbf{43.89}\\
         \hline
         11 & \textbf{0}& \textbf{0}& 90.39 & \textbf{0}& 1047 & \textbf{87.63}\\
         \hline
         12 & \textbf{0}& \textbf{11}& 163.26 & \textbf{0}& 1158 & \textbf{156.31}\\
         \hline
         13 & \textbf{0}& \textbf{46}& 358.42 & \textbf{0}& 1633& \textbf{340.08}\\
         \hline
         1 & \textbf{0}& \textbf{0}& 55.46 & 4& 3386& 32.15\\
         \hline
         7 & \textbf{0}& \textbf{0} & 100.66 & 5 & 3844 &69.96\\
         \hline
         8 & \textbf{0}& \textbf{3} & 173.65 & 5& 6837& 141.52\\
         \hline
         9 & \textbf{0}& \textbf{3} & 389.16 & 7& 20131 & 323.68\\
         \hline
    \end{tabular}
\end{table}

\subsection{Experiments on Physical Robots}

We test our algorithm in 2D using iRobot Create2s, Turtlebot3s, Turtlebot2s, and in 3D using Crazyflie 2.0s.
The details of our physical experiments are provided in the journal submission.



The recordings for our physical robot experiments are included in the supplemental video.

\section{Conclusion}

In this work, we present RLSS, an algorithm for distributed real-time trajectory replanning that i) considers dynamic limits of the robots explicitly, ii) enforces safety as hard constraints, iii) can work in real-time, iv) requires sensing only the positions of other agents and obstacles, v) does not use communication, and vi) empirically avoids deadlocks. 
Our approach enables robots to navigate in cluttered environments with high reactivity.
Our conservative requirements of only requiring to sense the positions of other robots and no explicit communication, increase the applicability of RLSS to a wide range of settings, including cases where velocity estimates are very noisy and communication is lossy or unavailable.

In empirical comparison to DMPC, our algorithm takes more time per iteration and computes slightly less distance-efficient solutions. However, DMPC sacrifices completeness and correctness: RLSS has considerably fewer collisions and avoids deadlocks effectively.
We show our algorithm's applicability to physical robots by demonstrating its behavior on Turtlebot2s, Turtlebot3s, iRobot Create2s, and Crazyflie 2.0s.

In future work, we would like to extend the algorithm with communication to improve the quality of the resulting trajectories. 
Also, in the future, we would like to incorporate the noise in the sensing systems within our algorithm to bring it closer to real world deployment.

\bibliographystyle{IEEEtran}
\bibliography{bibliography}

\end{document}